\documentclass{article}
\usepackage{spconf,amsmath,epsfig}
\usepackage{url}
\usepackage[caption=false]{subfig}
\usepackage{graphicx}
\usepackage{graphics}
\usepackage{multirow}
\begin{document}
\sloppy

\def\x{{\mathbf x}}
\def\L{{\cal L}}

\title{Multi-task deep CNN model for no-reference image quality assessment on smartphone camera photos}
\name{$^1$Chen-Hsiu Huang and $^2$Ja-Ling Wu}
\address{Department of Computer Science and Information Engineering,\\National Taiwan University, Taiwan\\
    E-mail: $^1$chenhsiu48@cmlab.csie.ntu.edu.tw, $^2$wjl@cmlab.csie.ntu.edu.tw}

\maketitle

\begin{abstract}
Smartphone is the most successful consumer electronic product in today's mobile social network era. The smartphone camera quality and its image post-processing capability is the dominant factor that impacts consumer's buying decision. However, the quality evaluation of photos taken from smartphones remains a labor-intensive work and relies on professional photographers and experts. As an extension of the prior CNN-based NR-IQA approach, we propose a multi-task deep CNN model with scene type detection as an auxiliary task. With the shared model parameters in the convolution layer, the learned feature maps could become more scene-relevant and enhance the performance. The evaluation result shows improved SROCC performance compared to traditional NR-IQA methods and single task CNN-based models.
\end{abstract}


\section{Introduction}
\label{sec:intro}

Image quality assessment (IQA) methods are developed to automatically to predict image quality without human subjective judgment, which is known to be costly and time-consuming. It was evident that various image distortions such as blur, noise, and JPEG compression artifact highly impact our perceived visual quality. Thus, plenty of distortion-oriented image quality databases have been constructed to support IQA researches, such as LIVE \cite{sheikh2006statistical}, CSIQ \cite{larson2010most}, and TID2013 \cite{ponomarenko2015image}. However, in today's mobile and social network era, most of the consumer photos are taken by smartphones with high resolution and compelling quality, stored on the device or sent to the cloud. For smartphone camera photos, usually they're original in the raw format without distortion, or the only distortion comes from smartphone camera's built-in Image Signal Processor (ISP). As the most successful consumer electronic product, the smartphone camera quality and its ISP post-processing capability is the dominant factor that impacts consumer's purchase decision. Currently, the evaluation of smartphone camera photo quality still relies on professional photographers and experts.

The IQA methods can be categorized into three types depending on whether the original reference image is involved, which are full-reference IQA (FR-IQA), reduced-reference IQA (RR-IQA), and no-reference IQA (NR-IQA) methods. Although NR-IQA is more challenging due to lack of the original image, it remains an important research area because there are some application scenarios where the reference image is not available or even not exist, like the smartphone camera photos. Therefore, how to apply existing NR-IQA researches to smartphone camera photos becomes an urgent industrial need. To bridge the gap, a new smartphone camera photo database \cite{zhu2020multiple}, denoted as SCPQD2020, is established to address image quality assessment issues on smartphone camera photos. Although there is no distortion in the collections of SCPQD2020, but the camera lens, the image sensors, and the processor decide the perceived quality.

\begin{figure*}[ht]
\begin{center}
\includegraphics[width=15cm]{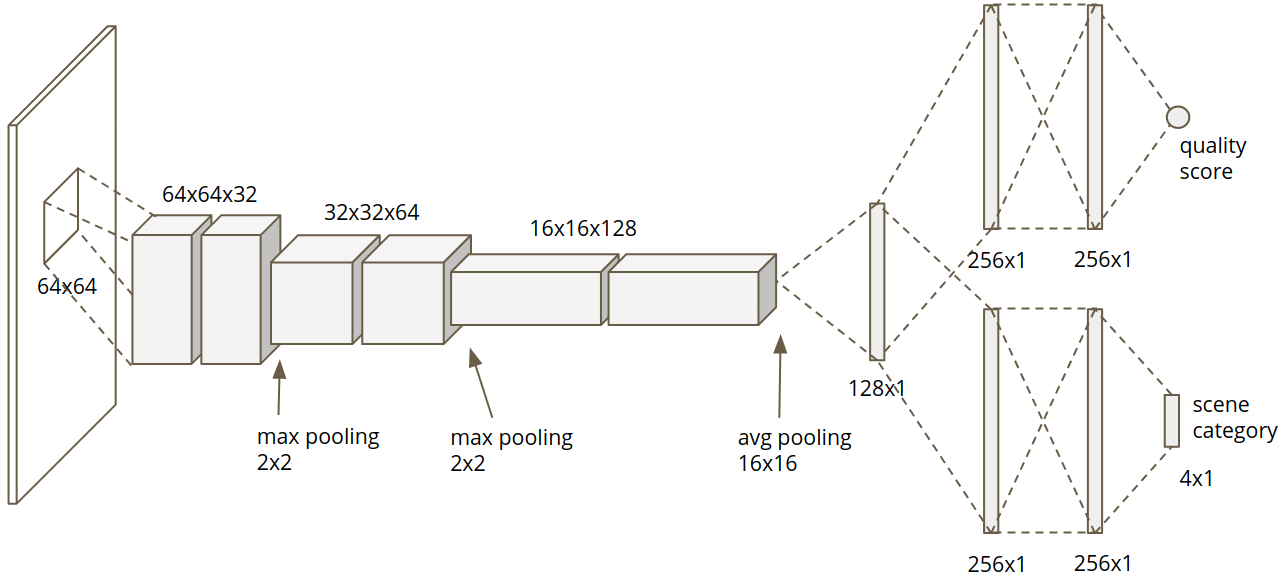}
\end{center}
\caption{The proposed multi-task CNN model with scene type as an auxiliary task.}
\label{fig:arch-dcnns}
\end{figure*}

In this work, a multi-task deep convolutional neural network (CNN) model is proposed for NR-IQA on smartphone camera photos, denoted as DCNNS. We add a secondary task, scene type detection as an auxiliary task to the primary quality prediction task, and simultaneously train the network with shared convolutional layer parameters. Through evaluation, we demonstrate that our multi-task CNN model does lead an optimization process to update the network weights for fitting the authentic distortion of smartphone photos and achieves better quality prediction.

This paper is organized as follows. We first review some relevant NR-IQA works in Section \ref{sec_related}. Section \ref{sec_propose} describes and discusses our proposed multi-task CNN model in detail. Then, we evaluate our DCNNS model and compare it to related NR-IQA methods, in Section \ref{sec_eval}. Finally, we conclude this writeup in the last section.

\section{Related Works} \label{sec_related}

The NR-IQA researches could be divided into two categories. One category focuses on designing better-handcrafted features, i.e., Natural Scene Statistics (NSS) features, characterizing the distribution of specific filter responses in the wavelet  \cite{moorthy2011blind} or DCT \cite{saad2012blind} transform domain, but this approach is too slow to be used in real-world applications. BRISQUE \cite{mittal2012no} and NIQE \cite{mittal2012making} are later developed to extract features from the spatial domain with reduced computational time. The other category focuses on feature learning, which attempts to learn discriminant visual features automatically without handcrafting. CORNIA \cite{ye2012unsupervised} demonstrates that it's possible to learn discriminant image features from raw image pixels, instead of using handcrafted features.

To learn discriminant features, the successfully proven Convolutional Neural Network (CNN) models in computer vision and image recognition did inspire NR-IQA researchers. Kang et al. \cite{kang2014convolutional} applied a CNN model on their NR-IQA work and achieve state-of-the-art performance on LIVE dataset with good cross-database generalization ability. Later the same CNN architecture was revised to a compact multi-task CNN for simultaneously estimating image quality and identifying distortion types \cite{kang2015simultaneous}. By reconsidering the quality prediction task as a multi-task problem of two different high-level tasks, the shared convolutional features help to achieve a similar or better performance of the state-of-the-art.

Deeper CNN model \cite{bosse2016neural} and fine-tuning pre-trained CNN model on large image datasets \cite{li2016no,kim2017deep} were employed to develop more accurate NR quality prediction methods. However, as Kim et al. pointed out in their survey \cite{kim2017deep}, a deeper or pre-trained CNN model increases the performance of NR-IQA to a competitive level of FR-IQA and handcrafted feature-based NR-IQA methods, but mainly on legacy synthesis distortion databases \cite{sheikh2006statistical, larson2010most, ponomarenko2015image}. The quality prediction result on authentic distortion databases, such as LIVE "In the Wild" Challenge Database \cite{ghadiyaram2015massive} and SCPQD2020 \cite{zhu2020multiple} are still far behind the accuracy of legacy databases.

As traditional NR-IQA methods show unsatisfying results \cite{zhu2020multiple} on smartphone photos, Yao et al. \cite{yao2020convolutional} proposed a CNN model with residual block and integrated the feature extraction and regression into one optimization process to predict image quality. They select the saliency regions using saliency maps generated by SalGAN, and carefully extract feature maps on different aspects such as HSV color space conversion and Gabor wavelet, then send them to CNN model as input. Their experiments show a better performance for smartphone images than traditional NR-IQA methods.

\section{Proposed Method} \label{sec_propose}

The proposed multi-task CNN model with scene type detection as an auxiliary task is presented in Figure \ref{fig:arch-dcnns}. As Kim et al. \cite{kim2017deep} mentioned in their survey, we adopt a deep CNN architecture for no-reference image quality assessment on smartphone camera photos to leverage CNN's strong representation and generalization capability. There are four quality measurement aspects, texture, color, noise, and exposure \cite{zhu2020multiple} in smartphone camera captured images. It's intuitive to exploit different image color spaces or extract low-level features with handcrafted filters like \cite{yao2020convolutional}. However, our attempts show no significant difference in a deeper CNN network, so we use the converted grayscale image directly. Like traditional NR-IQA methods \cite{mittal2012no}, we found that pre-processing can increase the performance and employed local contrast normalization as follows:

\[
\begin{aligned}
\hat{I}(i,j) & = \frac{I(i,j)-\mu(i,j)}{\sigma(i,j)+C} \\
\end{aligned}
\]

The normalized pixel $\hat{I}(i,j)$ at position $(i,j)$ is obtained from subtracting the original pixel value $I(i,j)$ by $\mu(i,j)$, which is the mean of pixel values within a local $P\times Q$ window centered at $(i,j)$. Then, we divided the result by $\sigma(i,j)$, which is the standard deviation of pixel values within the window. The constant $C=1$ is used to prevent divided by zero. We calculate $\mu(i,j)$ and $\sigma(i,j)$ with setting $P=Q=3$ as follows:

\[
\begin{aligned}
\mu(i,j) & = \sum_{p=-P}^P \sum_{q=-Q}^Q I(i+p,j+q)\\
\sigma(i,j) & = \sqrt{\sum_{p=-P}^P \sum_{q=-Q}^Q (I(i+p,j+q) - \mu(i, j))^2}
\end{aligned}
\]

We crop the input images into non-overlapping $64\times64$ patches with stride 160. Different sizes of stride could be used, but since the original smartphone camera captured images are in high resolution and we choose relatively small-sized patches, the resulting segmentation gives us enough training data to learn.

Through various experiments, we found that CNN can effectively iterate and update the learned model parameters in the loss minimization process, but also easy to be trapped without gaining further improvement if a certain level of performance is reached. As multi-task learning in neural networks demonstrated that learning multiple correlated tasks at the same time may improve the overall performances \cite{caruana1997multitask}, we reformulate the quality prediction problem as a multi-task problem of two different high-level tasks. We add a scene type detection auxiliary task as a secondary task to train simultaneously with the quality score prediction task in one CNN. We denote our proposed multi-task CNN model as DCNNS and illustrate details of our method in the following sections.

\subsection{Deep Multi-task CNN Model}

The convolution layer of proposed DCNNS adopts a similar but smaller deep CNN architecture like VGG-16 \cite{simonyan2014very}. The six convolution layers all use $3\times3$ filter size with stride as 1 and padding by 1, which lead to two $64\times64\times32$ layers, two $32\times32\times64$ layers, and two $16\times16\times128$ layers. Max pooling of $2\times2$ is performed twice to reduce the feature map resolution from $64\times 64$ to $16\times 16 $. After the final convolution layer, apply $16\times16$ global average pooling to flatten feature maps to a 128-dimension fully connected layer as output.

In the quality score prediction task, the convolution output layer connects to two 256 fully connected (FC) layers, then regress to an overall quality score. The scene detection task has the same structure as that of the quality task, except for the last layer, which is composed of four neurons to represent the probabilities of different scene types. Dropout of probability 0.5 is applied to the two 256 FC layers to prevent overfitting. All the layers use ReLU as activation functions.

Because the two tasks share the same model parameters in the convolution layer, the learned feature maps could become more scene-relevant and better model the image quality.

\subsection{Scene Categorization}

We observed that there are typical types of scenes in smartphones taken photos, such as outdoor nature scenes, daylight buildings, indoor facilities, and night scenes, etc. Different scenes have different characteristics and affect human's perceptual quality judgment. It is essential to provide the scene information as a clue for quality prediction during training. For categorizing scene types, images are horizontally and vertically divided by 8, forming 64 sub-blocks, then mean and standard derivation of each sub-block is calculated and split into 16 bins histogram, accumulating through all sub-blocks. To detection different kinds of edges, we apply edge filters of MPEG-7 image descriptors \cite{manjunath2001color}, as shown in Figure \ref{fig:edge-filter} on each sub-block and calculate the percentage of edge pixels, then accumulate across all sub-blocks. As a result, we extract a 37-dimension feature vector for each image and utilize the unsupervised K-means algorithm to cluster images into four types of scenes.

\begin{figure}[!ht]
\begin{gather*}
\begin{bmatrix}
1 & -1 \\
1 & -1
\end{bmatrix}
\begin{bmatrix}
1 & 1 \\
-1 & -1
\end{bmatrix}
\begin{bmatrix}
\sqrt{2} & 0 \\
0 & -\sqrt{2}
\end{bmatrix} \\
\begin{bmatrix}
0 & \sqrt{2} \\
-\sqrt{2} & 0
\end{bmatrix}
\begin{bmatrix}
2 & -2 \\
-2 & 2
\end{bmatrix}
\end{gather*}
\caption{Edge filters for finding scene clusters.}%
\label{fig:edge-filter}%
\end{figure}

\begin{figure}[htb]
\centering
\subfloat[Scene 0: night scenes]{{\includegraphics[width=3.5cm]{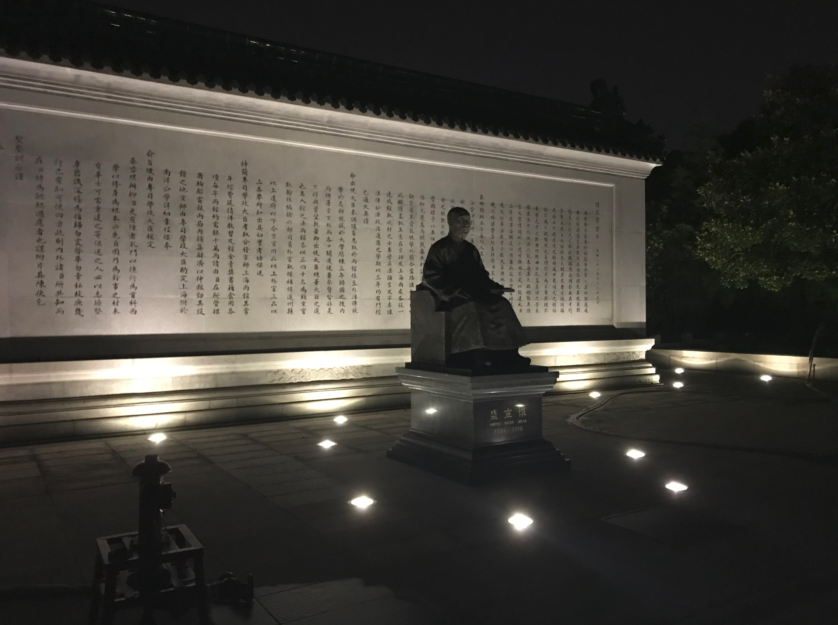} }} \quad
\subfloat[Scene 1: scenes with complex texture and balanced exposure]{{\includegraphics[width=3.5cm]{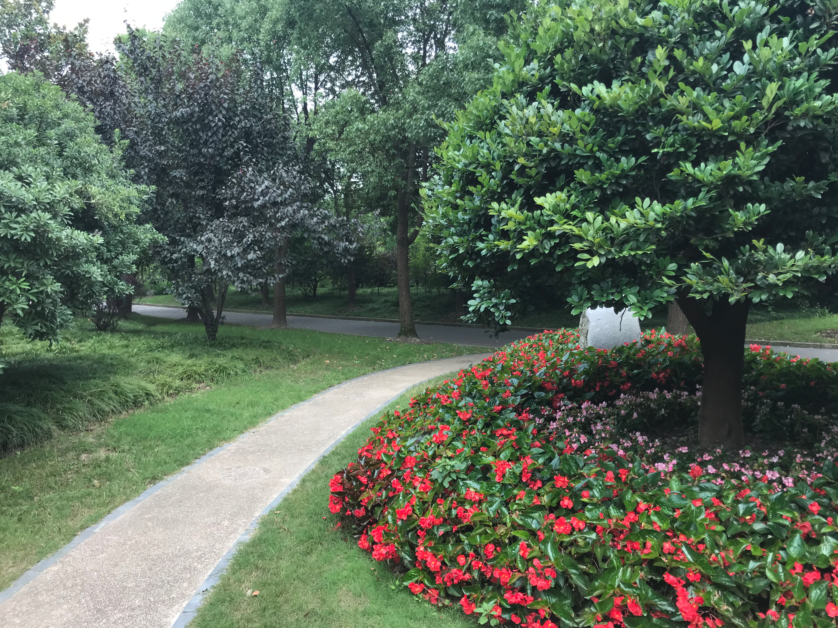} }} \\
\subfloat[Scene 2: scenes with balanced lighting and contain both smooth region and complex texture]{{\includegraphics[width=3.5cm]{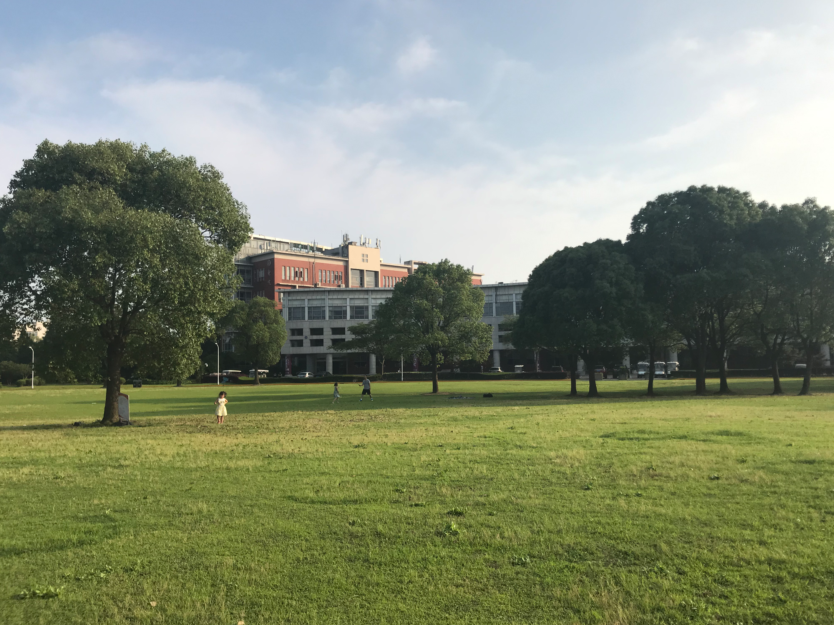} }} \quad
\subfloat[Scene 3: scenes with less contrast and under exposure]{{\includegraphics[width=3.5cm]{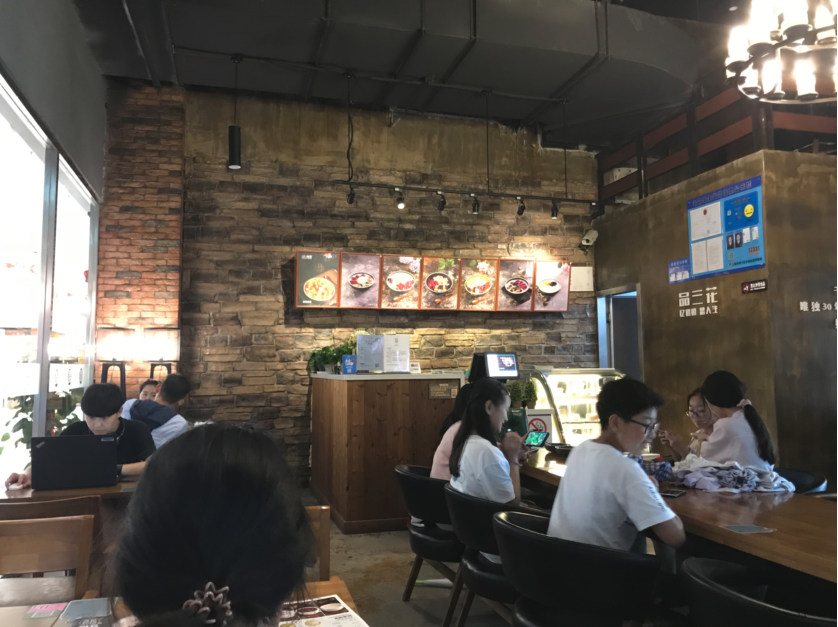} }}
\caption{Sample images of scene categorization result}
\label{fig:scene-cat}%
\end{figure}

The clustering result provides the scene type label for the scene detection task. Figure \ref{fig:scene-cat} shows some sample images of the categorization result. We can see that essential photography characteristics such as contrast, brightness, and texture are distinguishable in each scene type.

\subsection{Loss Function and Learning}

Let $x_n$ and $y_n$ denote the input patch and its quality label. $f(x_n;w)$ is the function of $x_n$ and network weights $w$ that predicts the quality score. The loss function of quality task $Lq$ is defined as follows:

\begin{equation}
L_q = \frac{1}{N}\sum_{n=1}^{N} \lVert f(x_n;w)-y_n \lVert_{l_1}
\end{equation}

We apply softmax function on the probability of scene category and adopt cross-entropy as loss function $L_s$, where $f'(x_n;w')$ is the function of $x_n$ and network weights $w'$ that predicts scene type. Network weights $w$ and $w'$ share the same convolution layer parameters except for FC layers. We define $L_s$ as follows, where $y_n^{(s)}$ denotes the scene label of $x_n$ from clustering:

\begin{equation} \label{eq:Ls}
L_s = \frac{1}{N} \sum_{n=1}^{N} H(f'(x_n;w'),y_n^{(s)})
\end{equation}

\begin{figure*}[ht]
\begin{center}
\includegraphics[width=17.5cm]{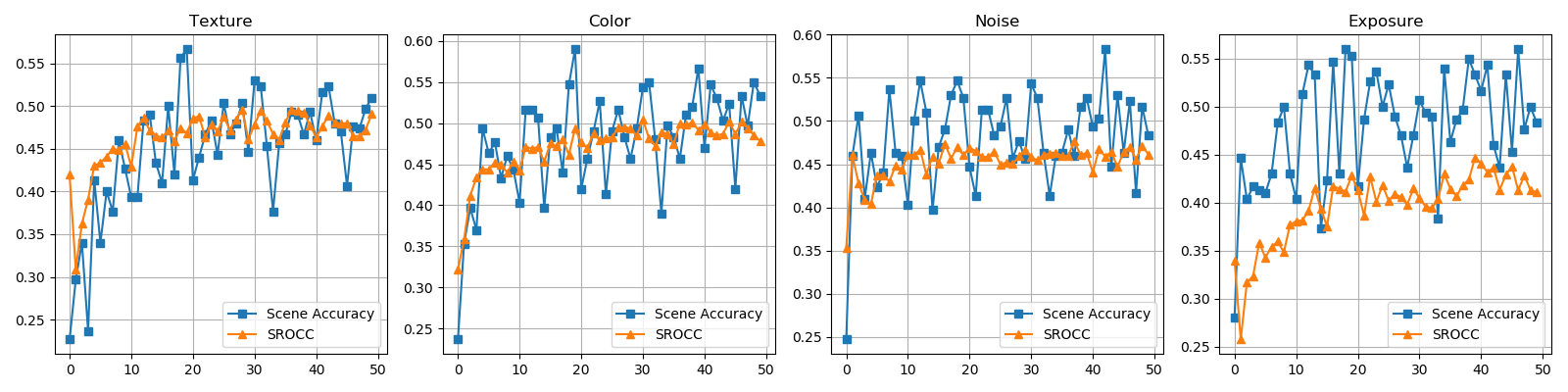}
\end{center}
\caption{The validation trending of SROCC and scene type detection accuracy for each epoch.}
\label{fig:train-acc}
\end{figure*}

The overall loss function is defined as $L=L_q+\alpha L_s$. We weight the scene detection loss function $L_s$ by $\alpha=1.0$ to guide the training process to learn more scene-relevant features. When we train the two tasks simultaneously with the combined loss function $L$, the shared convolution layer parameters are updated in a way that improves the quality prediction and scene detection accuracy at the same time, as shown in Figure \ref{fig:train-acc}. Each scene type's detection accuracy varies under four quality aspects, ranging from 0.28 to 0.58 and depending on different scene's characteristics. Although the averaged scene type detection accuracy is around 0.41, the auxiliary task can be thought as a weak classifier to boost the performance. Table \ref{tab:scene-acc} shows the scene type detection task's accuracy for each quality aspect at the corresponding best epoch.

\begin{table}[ht]
\centering
\caption{Scene type detection task accuracy}
\label{tab:scene-acc}
\begin{tabular}{rrrrr} \hline
\multicolumn{1}{l}{Scene} & \multicolumn{1}{l}{Texture} & \multicolumn{1}{l}{Color} & \multicolumn{1}{l}{Noise} & \multicolumn{1}{l}{Exposure} \\ \hline
\multicolumn{1}{c}{0} & 0.2867 & 0.3977 & 0.4103 & 0.2836 \\
\multicolumn{1}{c}{1} & 0.4228 & 0.3428 & 0.2852 & 0.4371 \\
\multicolumn{1}{c}{2} & 0.3629 & 0.4131 & 0.3730 & 0.4095 \\
\multicolumn{1}{c}{3} & 0.5639 & 0.5740 & 0.5869 & 0.5285 \\ \hline
\multicolumn{1}{c}{Average} & 0.4091 & 0.4319 & 0.4138 & 0.4147 \\ \hline
\end{tabular}
\end{table}

For training, we split the SCPQD2020 database into 80\%-20\% parts as training and validation sets. We employ Adam optimizer with default settings in PyTorch to optimize our network parameters with learning rate 0.001 and batch size 128. We train one CNN model for each quality aspect with 50 epochs. Each model spends about 1.5 hours to complete the 50 epochs on an NVIDIA GeForce RTX 2080 Ti GPU with 11GB RAM.

\section{Experimental Results} \label{sec_eval}

We compare our proposed multi-task CNN-based NR-IQA method with traditional NR-IQA methods BRISQUE \cite{mittal2012no} and NIQE \cite{mittal2012making}. As expected, the traditional NR-IQA methods do not perform well on smartphone camera photos. We also compare our work with two CNN-based IQA methods, one is a shallow CNN network \cite{kang2014convolutional} denoted as CNNIQA, and another is a CNN with residual block approach proposed by Yao et al \cite{yao2020convolutional}.

\subsection{Dataset}

The SCPQD2020 dataset \cite{zhu2020multiple} is composed of 1,500 photos taken from 100 scenes using 15 smartphones, covering a wide range of prices and different manufacturers. The dataset includes various challenging scenes like outdoor nature scenes, indoor lowlight scenes, backlight scenes, and night scenes. For every image of the same scene, three professional photographers are recruited to rate subjective quality scores based on four aspects: texture, color, noise, and exposure.
The exact quality scores are not provided in the dataset but the quality ranking of 15 images within a scene is disclosed, ranging from 1 to 15. Lower-ranking means better quality.

\subsection{Performance Metric}

The quality prediction result of each quality aspect is sorted and compared with ground truth subjective ranking using the Spearman Rank Order Correlation Coefficient (SROCC). SROCC mainly reflects the consistency between two ranking distributions and is defined as:

\[
SROCC = 1- \frac{6\sum_i d_i^2}{n(n^2-1)}
\]

where $n$ is the number of images and $d_i$ is the rank difference between the subjective score and the objective prediction of the $i$-th image.

\subsection{Performance Comparison}

We randomly split the dataset into 80\%-20\% parts as the training and the testing sets. Each quality aspect has a separate prediction model, and we train our model with 50 epochs and select the model with the highest SROCC, repeating the processes three times to find the average score. We use the same training sets to investigate the competing NR-IQA methods BRISQUE, NIQE, and CNNIQA to report their averaged SROCC values. Since the work from Yao et al. is not publicly available, we quote the performance number from their paper \cite{yao2020convolutional} as a reference. Table \ref{tab:perf-res} shows the SROCC comparison.

\begin{table}[htb]
\centering
\caption{SROCC comparison of the proposed method and other IQA methods. The superior result is marked in bold.}
\label{tab:perf-res}
\begin{tabular}{ccccc}
\hline
Model & Texture & Color & Noise & Exposure \\ \hline
BRISQUE \cite{mittal2012no} & 0.2065 & 0.1973 & 0.2139 & 0.2175 \\
NIQE \cite{mittal2012making} & 0.2518 & 0.2446 & 0.1896 & 0.2459 \\ \hline
CNNIQA \cite{kang2014convolutional} & 0.4405 & 0.4657 & 0.4485  & 0.3793 \\
Yao et al. \cite{yao2020convolutional} & 0.4414 & 0.4827 & 0.4525 & \textbf{0.4368} \\
DCNNS & \textbf{0.4899} & \textbf{0.4979} & \textbf{0.4723} & 0.4349 \\ \hline
\end{tabular}
\end{table}

As we expected, the traditional NR-IQA methods do not perform well since they're designed for synthesis distortions not for smartphone camera photos which are original raw images without distortion. When compared with a CNN-based NR-IQA model, CNNIQA, our proposed DCNNS model excels at both deeper network architecture and the scene auxiliary task that assists to learn more scene-relevant features. Although it may not be a head-to-head comparison, we quote the performance result from Yao to indicate that a deeper network with multi-task learning can achieve similar or superior performance, as their approach is compound with saliency detection, feature extraction, and color space conversion.

\subsection{Discussion} \label{sec_scene_det}

\begin{figure}[htb]
\centering
\subfloat[Image 55 is a night scene (scene 0) but clustered as scene 2, balanced lighting with smooth and complex. This image has lowest detection accuracy 0.03] {{\includegraphics[width=4cm]{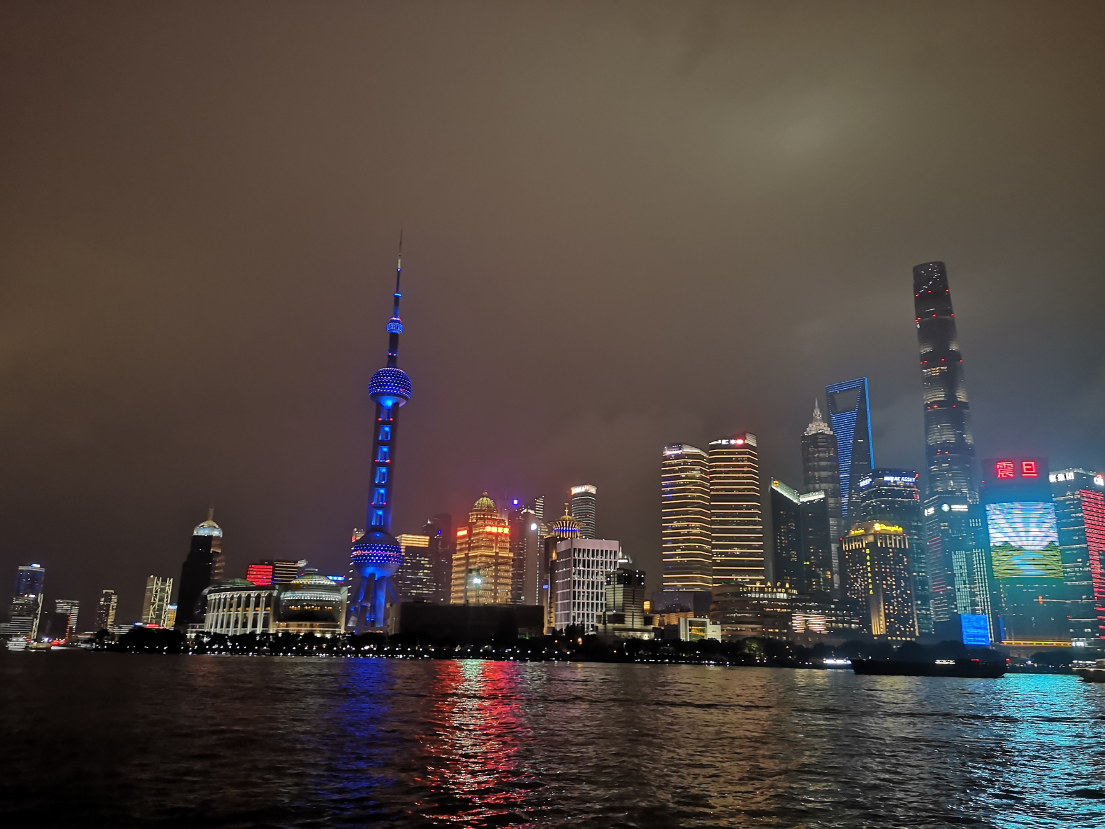} }} \\
\subfloat[Patches from specific region are misclassified as scene 3, less contrast and under exposure]  {{\includegraphics[height=3cm]{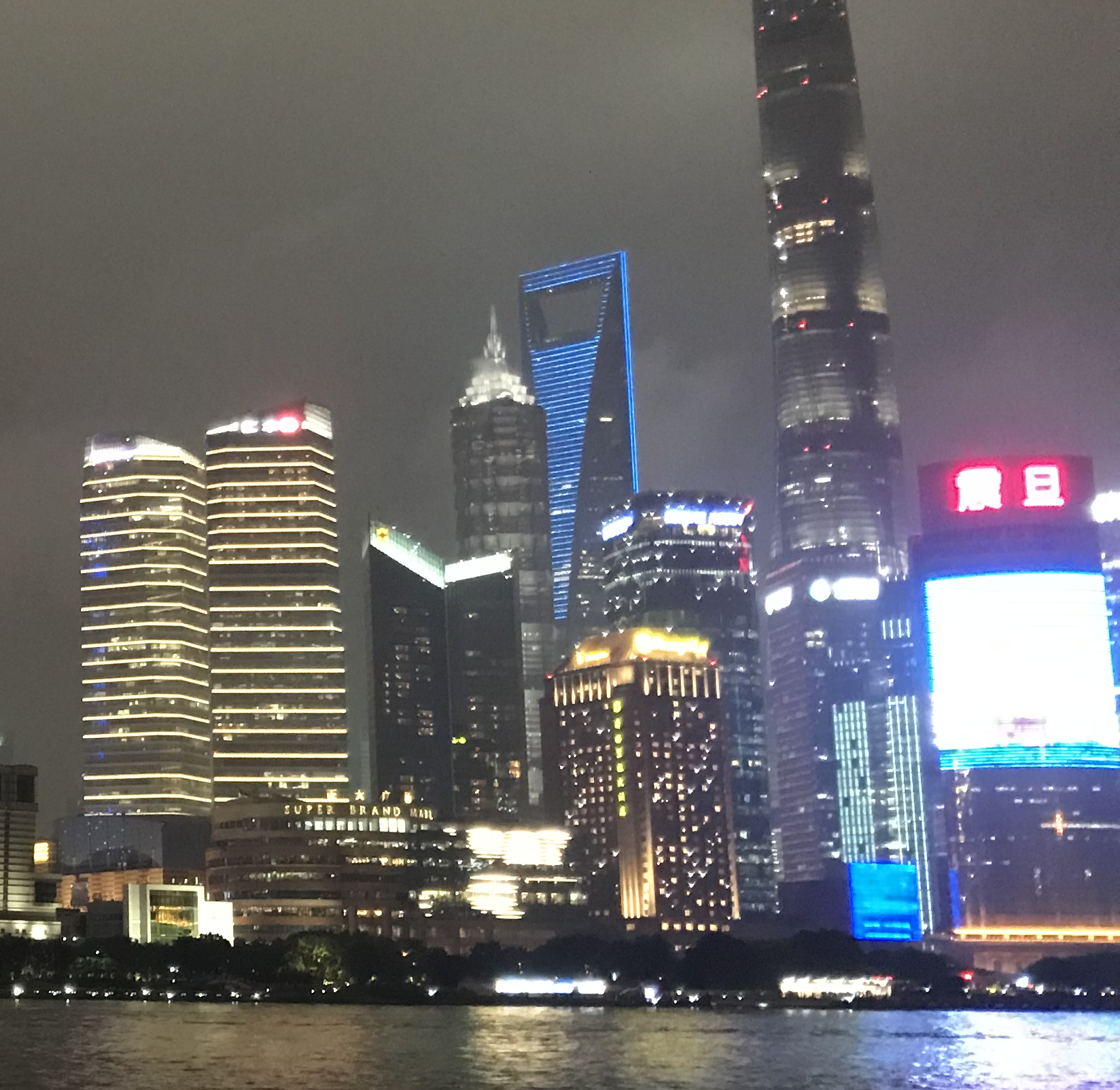} } \label{fig:p55_hi}} \quad
\subfloat[Patches from specific region are misclassified as scene 0, night scene] {{\includegraphics[height=3cm]{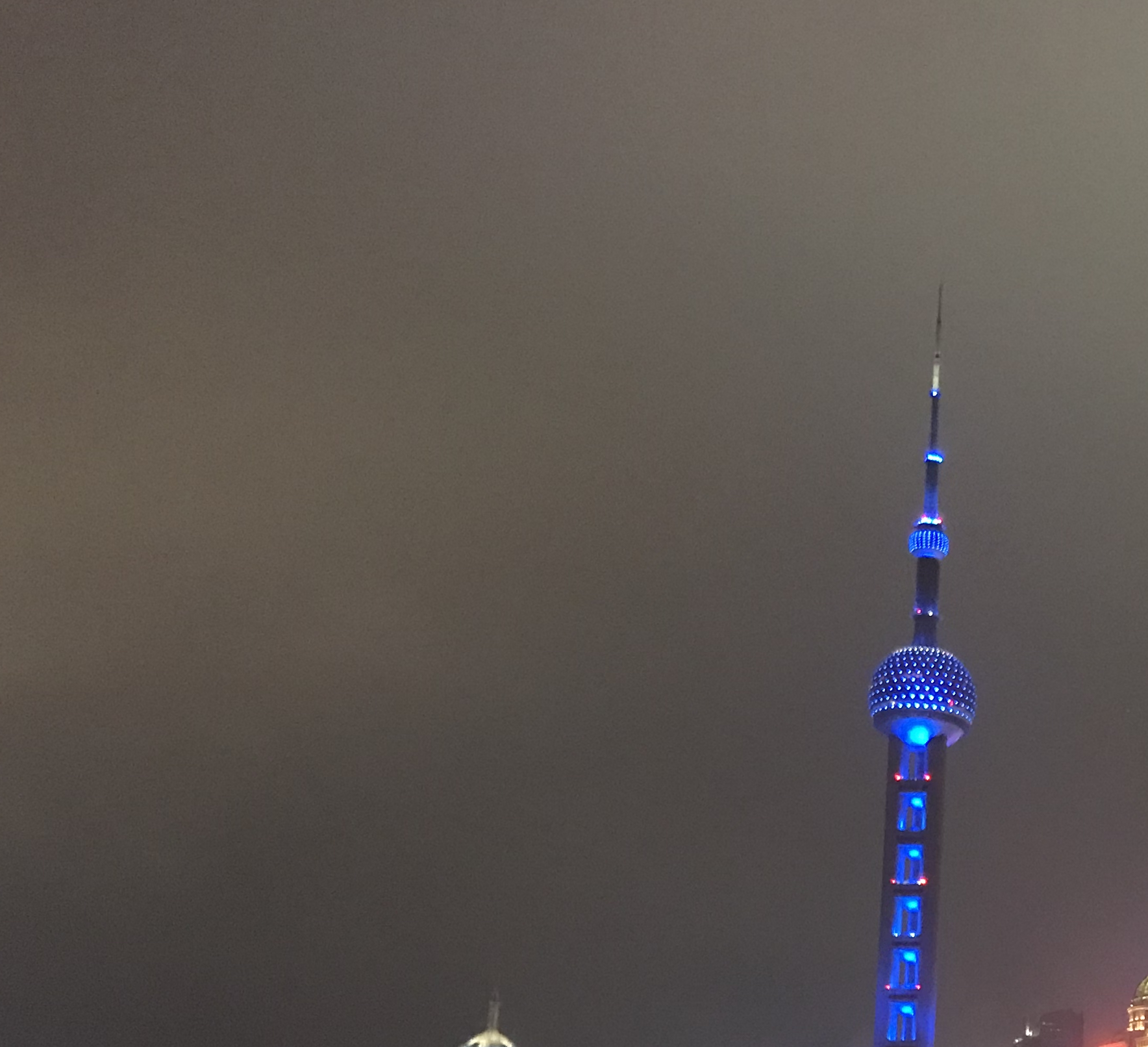} } \label{fig:p55_lo}}
\caption{Night scene images with low scene accuracy}
\label{fig:night-scene}%
\end{figure}

Although we use scene type detection as an auxiliary task to assist more accurate image quality prediction, the scene type concept comes globally from all image parts and sometimes from image context. As shown in Figure \ref{fig:night-scene}, the image 55 is a night scene from its context, but clustered as scene 2 image-wise, given it's the scene with balanced lighting and contains both smooth and complex region.

Since we label the scene type and predict the image quality on image patches of size $64\times 64$, the scene type identified globally may not fit to each individual patch from different image regions. During the training process, some patches from specific region (Figure \ref{fig:p55_hi}) are classified as scene 3, the scene with less contrast and under exposure, while the other region (Figure \ref{fig:p55_lo}) are classified as scene 0, the night scene. The labelled ground truth scene 2 violates what the neural network learned from other training samples and keep penalizes the optimization process with loss function $L_s$ in equation (\ref{eq:Ls}). As a result, the image 55 has the lowest average scene detection accuracy for every quality aspect and lower SROCC scores. We present some representative images' SROCC scores and scene detection accuracies in Table \ref{tab:image-ana}.

\begin{table}[ht]
\centering
\caption{SROCC v.s. Scene accuracy for selected images}
\label{tab:image-ana}
\resizebox{\columnwidth}{!}{%
\begin{tabular}{clrrrr} \hline
Image &  & \multicolumn{1}{l}{Texture} & \multicolumn{1}{l}{Color} & \multicolumn{1}{l}{Noise} & \multicolumn{1}{l}{Exposure} \\ \hline
\multirow{2}{*}{55} & SROCC & 0.4286 & 0.2950 & -0.1827 & 0.2631 \\
 & Scene Acc. & 0.0289 & 0.0323 & 0.0326 & 0.0324 \\ \hline
\multirow{2}{*}{85} & SROCC & 0.5869 & 0.3874 & 0.3805 & 0.6970 \\
 & Scene Acc. & 0.7785 & 0.7028 & 0.6022 & 0.7750 \\ \hline
\multirow{2}{*}{47} & SROCC & 0.6200 & 0.7296 & 0.7621 & 0.3806 \\
 & Scene Acc. & 0.6714 & 0.6746 & 0.6750 & 0.6777 \\ \hline
\end{tabular}%
}
\end{table}

\begin{figure}[htb]
\centering
\subfloat[Image 85: scene 1] {{\includegraphics[width=3.5cm]{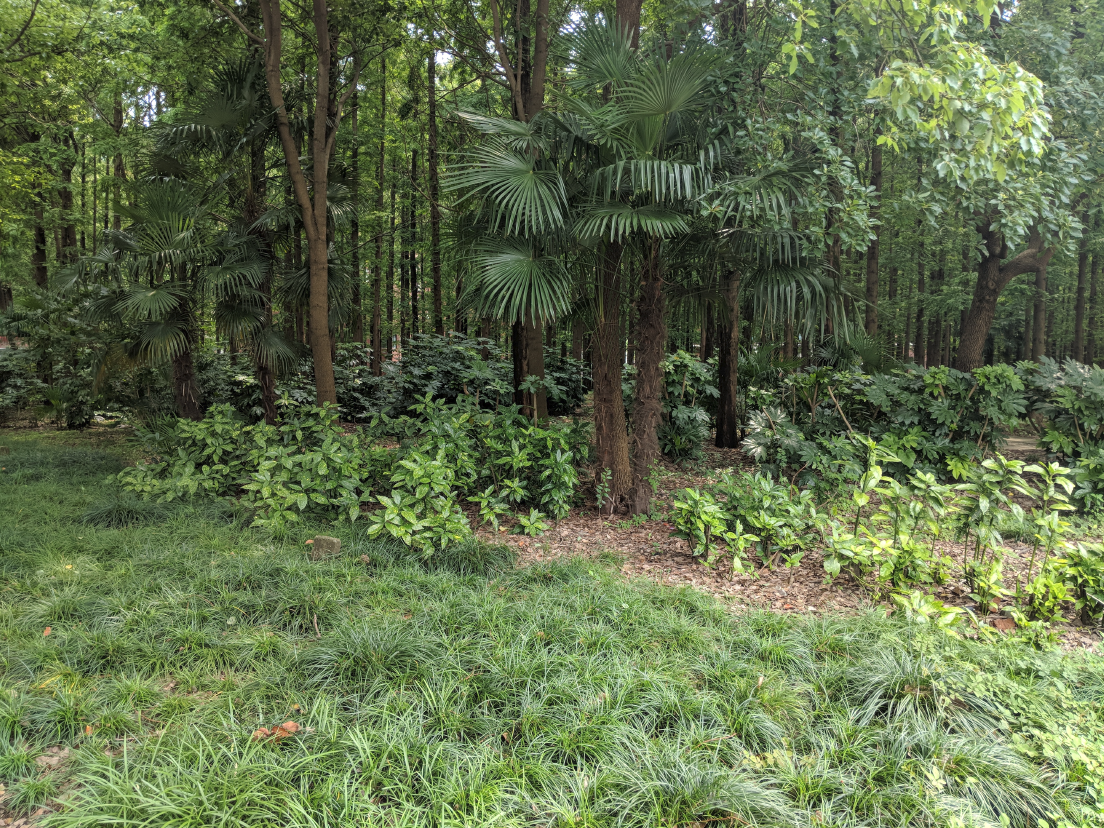} }} \quad
\subfloat[Image 47: scene 3] {{\includegraphics[width=3.5cm]{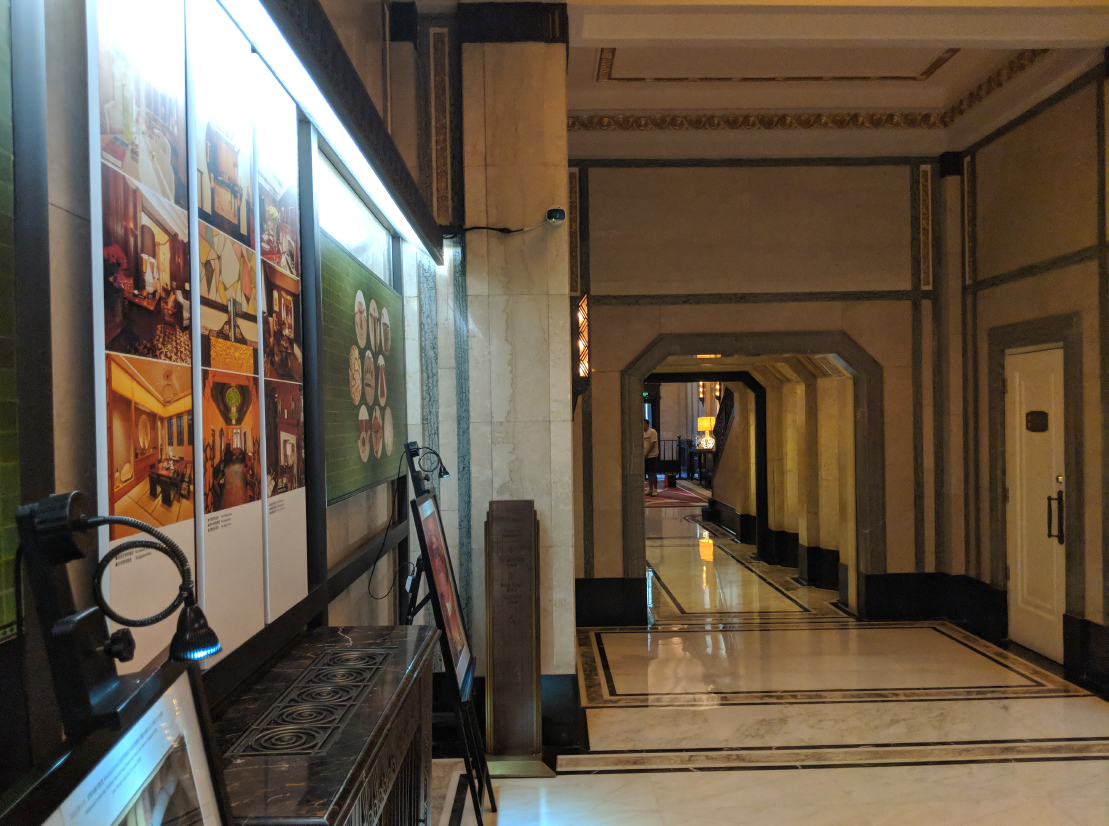} }} \\
\caption{Images with higher scene accuracy have higher than average SROCC}
\label{fig:image-other}%
\end{figure}

On the other hand, we investigate some images with overall high scene accuracy. From Figure \ref{fig:image-other}, the image 85 is labelled as scene 1, the scene with complex texture and balanced exposure, achieves average scene accuracy 0.7146 and results to above average SROCC as 0.5130. Another image 47 of scene 3, scenes with less contrast and under exposure, also has high scene accuracy as 0.6747, which leads to significant higher SROCC 0.6231 than average. Detailed performance numbers are listed in Table \ref{tab:image-ana}

From the analysis of SROCC v.s. scene accuracy, we demonstrated that the scene accuracy does have high impact on the image quality prediction, which also echos the performance boost from the help of auxiliary task. As we pointed out the night scene detection challenge, how to effectively fuse the global image-wise feature v.s. the local patch-wise feature to correctly identify scene type will be a key to further improve smartphone photo quality assessment.

\section{Conclusion}

This paper proposes a multi-task deep CNN model with scene type detection as auxiliary task. A simple image clustering method is used to label scenes of images for the scene type detection task, therefore guide the optimization process to better fit the smartphone camera photos. The evaluation result shows improved SROCC performance compared to traditional NR-IQA methods and single task CNN-based models.

\begin{acknowledgement}
This work is partially supported by the Ministry of Science and Technology, Taiwan and CITI SINICA, ROC, under the grant numbers of MOST 108-2218-e-002-055, MOST 108-2221-E-002-103-my3, and Sinica 3012-C3447.
\end{acknowledgement}

\bibliographystyle{IEEEbib}

\end{document}